\documentclass[acmsmall]{acmart}
\usepackage{graphicx}  
\usepackage{fancyhdr}
\AtBeginDocument{%
  }

\setcopyright{acmlicensed}
\acmJournal{PACMCGIT}
\acmYear{2025} \acmVolume{8} \acmNumber{3} \acmArticle{37} \acmMonth{8}\acmDOI{10.1145/3736789}

\received{2025-02-01}
\received[revised]{2025-03-15}
\received[accepted]{2025-04-10}

\begin{document}
\citestyle{acmauthoryear}
\title{Multi Layered Autonomy and AI Ecologies in Robotic Art Installations}

\author{Baoyang Chen}
\orcid{0009-0009-4882-6788}
\email{baoyang.chen@gmail.com}
\affiliation{%
  \institution{Hong Kong University of Science and Technology (HKUST)}
  \city{Hong Kong SAR}
  \country{China}
}
\affiliation{%
  \institution{Central Academy of Fine Arts}
  \city{Beijing}
  \country{China}
}

\author{Xian Xu}
\orcid{0000-0002-2636-7498}
\email{xianxu@ust.hk}
\affiliation{%
  \institution{Hong Kong University of Science and Technology (HKUST)}
  \city{Hong Kong SAR}
  \country{China}
}

\author{Huamin Qu}
\authornote{Corresponding author}
\orcid{0000-0002-3344-9694}
\email{huamin@ust.hk}
\affiliation{%
  \institution{Hong Kong University of Science and Technology (HKUST)}
  \city{Hong Kong SAR}
  \country{China}
}


\begin{abstract}
This paper presents \textit{Symbiosis of Agents}, a large-scale installation by artist Baoyang Chen, integrating AI-driven robotic agents within an immersive, reflective environment, foregrounding the delicate balance between machine agency and artist authorship by harnessing emergent behaviors from self-organized AI robotic ecologies as a creative apparatus. This project draws upon the traditions of early cybernetics, rule-based conceptual art, and pioneering robotic art installations, featuring dynamic interactions among robotic arms, quadruped robots, their environment, and the audience. A sophisticated three-tiered “faith system” guides these agents, manifesting at the micro-level through adaptive strategies, at the mesoscopic level via narrative-driven “drives,” and at the macro level with an overarching directive. This architecture enables organic evolution of the robots’ behaviors in response to environmental cues and subtle audience presence, transforming passive spectators into active participants. This  project situates itself within a speculative terraforming scenario, against the backdrop of historical and ethical analogies—particularly the exploitation of marginalized laborers—the installation raises critical questions about responsibility in AI-driven societies. Through carefully orchestrated robotic choreography, AI-generated conceptual scripts, and atmospheric effects like responsive lighting and fog, \textit{Symbiosis of Agents} situates AI not merely as computational tool but as collaborator in a dynamic, evolving creative process. Exhibited internationally, it exemplifies how cybernetic principles, robotic art experimentation, and conceptual-art rule-making can converge, challenging prevailing notions of agency and authorship. By interrogating ethical considerations and exposing emergent storytelling, the work expands possibilities for human–machine collaboration in the ever-evolving domain of contemporary art.

\end{abstract}



\begin{teaserfigure}
  \centering
  \includegraphics[width=0.65\linewidth]{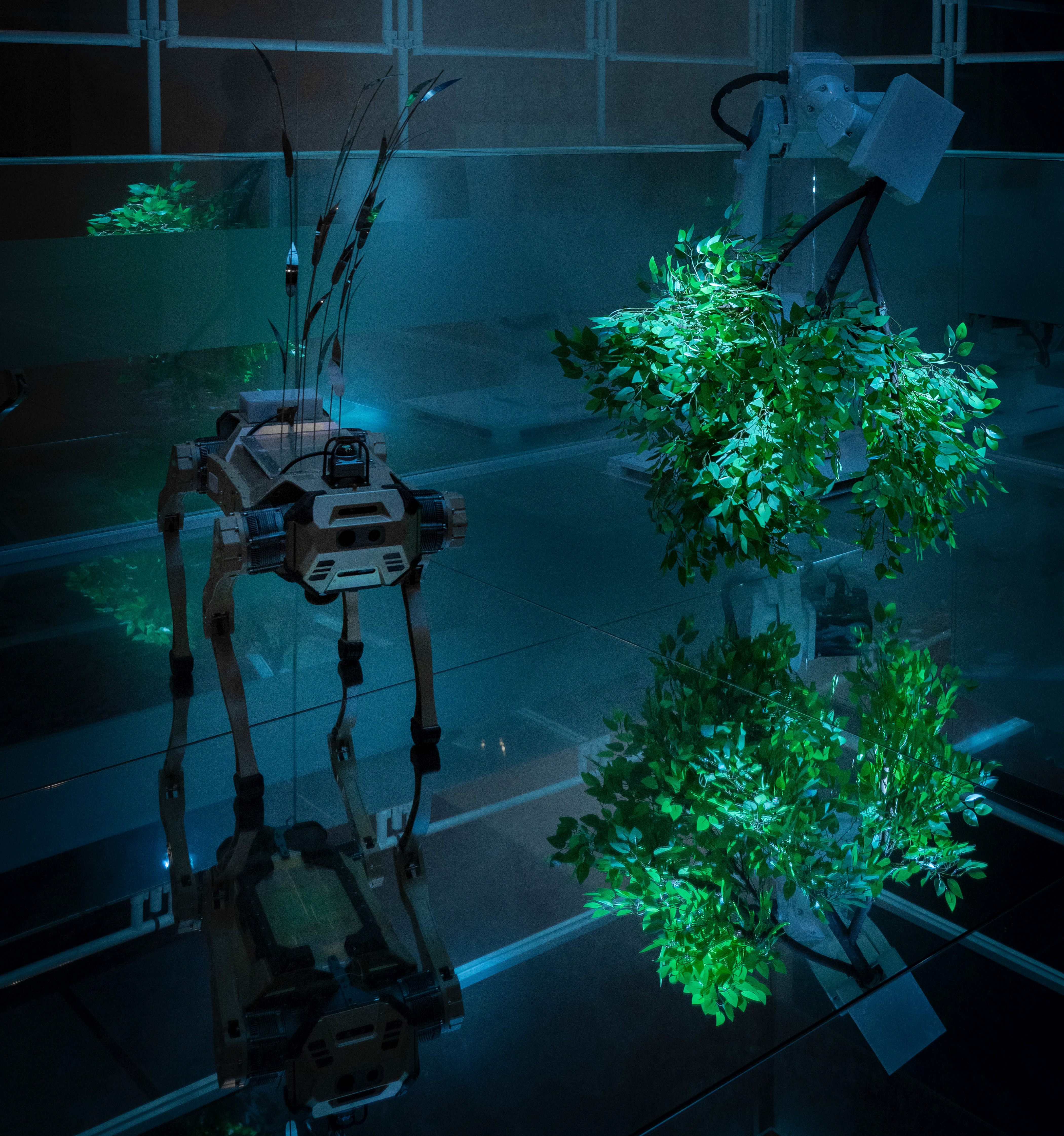}
  \Description{Night-lit installation scene: a quadruped robot walks on a mirror-finish floor, in front of a robotic arm that cradles a lush leafy branch; blue atmospheric lighting and soft fog create twin reflections of both robot and foliage on the glassy surface.}
  \caption{Installation View of the Artwork. ©Baoyang Chen}
  \label{fig:teaser}
\end{teaserfigure}


\maketitle
\renewcommand\refname{References}
\section{Introduction}
AI has evolved from generating screen‑based imagery to steering embodied
agents in artistic contexts \citep{Jeon2017,Sandoval2022}, spurring what
\citet{Kac2001} describes as “robotic art”—artworks that deploy autonomous
or semi‑autonomous machines in real‑world environments.  
This paper examines how multiple AI‑driven robotic agents can collectively
shape a dynamic narrative environment. It draws on two main traditions:
first, the legacy of conceptual and rule‑based artistic creation exemplified
by Sol LeWitt and Casey Reas, and second, the pioneering robotics‑based
practices of Stelarc, Goldberg, Moura, Donnarumma, and others who explore
performance, embodiment, and emergent expression in art‑making machines
\citep{Kac2001,HerathKroos2016}.

Paralleling these developments, early cybernetic artworks laid the
foundations for real‑time feedback and environmental adaptation.
Schöffer’s \textit{CYSP 1} (1956) and Ihnatowicz’s \textit{Senster} (1970)
integrated sensing, motors, and computation to produce lifelike reactions.
Gordon Pask’s cybernetic sculptures similarly explored conversational and
adaptive feedback \citep{Pask1968}, emphasizing that art‑making machines
could “listen” to their environment.  
Around the same period, Nicholas Negroponte’s \textit{SEEK} (1970)
proposed a robotic ecology of blocks and gerbils, and works such as
Goldberg’s \textit{The Telegarden} and Moura’s Swarm Paintings introduced
themes of emergent order that now undergird contemporary multi‑agent
systems in art.  
These historical precedents inform ongoing debates about where agency,
authorship, and creativity reside when the artwork itself can move, sense,
and respond—or even “learn” \citep{Bidgoli2020,Qin2025}.

In the past decade, researchers have noted how robotic installations can
reframe audience participation and authorial control
\citep{Mendez2019,Kumaran2024}, shifting from hands‑on user input toward
autonomous or adaptive narrative flows. This paper contributes to these
debates by emphasizing what \citet{Penny2013} calls “situated
machines”—embodied agents designed not for direct manipulation but for
observational and interpretive encounters. Drawing on immersive‑theatre
theory \citep{Simecek2024}, the installation discussed here prompts
viewers to develop reflexive self‑awareness by watching robots whose
behaviours arise from an AI‑driven “faith system” rather than external
guidance. This resonates with \citet{Vass2021}, who warn that adaptive
ecosystems can inadvertently introduce new forms of control, as well as
with \citet{Tresset2013}, whose performative drawing robots highlight
emergent expressions. Rather than rely on continuous human intervention,
each robotic arm or quadruped follows a textual policy, generating a
choreographed yet evolving interplay of lights, screens, and environmental
cues.

This multi‑agent installation invokes the frameworks of conceptual art,
engages the exploratory potential of robotic art, and employs narrative
design principles to foreground emergent storytelling. By merging
autonomous AI processes with performative spectacle, it aims to broaden
our understanding of how people encounter and critically reflect on
embodied AI in artistic settings—an enduring question since the earliest
cybernetic sculptures and telepresence installations
\citep{Goldberg1995,XSpace1993}. Through a new synthesis of action,
feedback, and viewer observation, this project expands on the notion of
interactivity, raising pressing questions about agency, perception, and
the future of human–machine relations in creative domains.

\section{MOTIVATION AND CONCEPT}

Building on the foundational ideas introduced earlier, \textbf{Symbiosis of Agents} created by artist Baoyang Chen, transitions AI-driven art from screen-based generative output (AIGC) into a more spatially grounded, agent-based framework. Physical robots, guided by policies derived from large language models, move autonomously within the gallery space and blend seamlessly into the architectural environment. This setup creates an immediate sense of presence, encouraging heightened observation and subtle, shared interactivity between viewers and the agents in the same physical setting. It aligns with broader trends in media arts, where cutting-edge technologies—particularly AI and robotics—open up new avenues for participation and immersion. Historically, the media arts field has expanded by embracing tools that span from kinetic sculptures to sophisticated digital installations. In \textbf{Symbiosis of Agents}, the robots operate as AI-driven entities occupying a narrative space, creating a self-evolving environment that transforms in real time, uniting classical principles of art viewing with the latest frontiers of technology.

\begin{figure}[ht!]
  \centering
  \includegraphics[width=\linewidth]{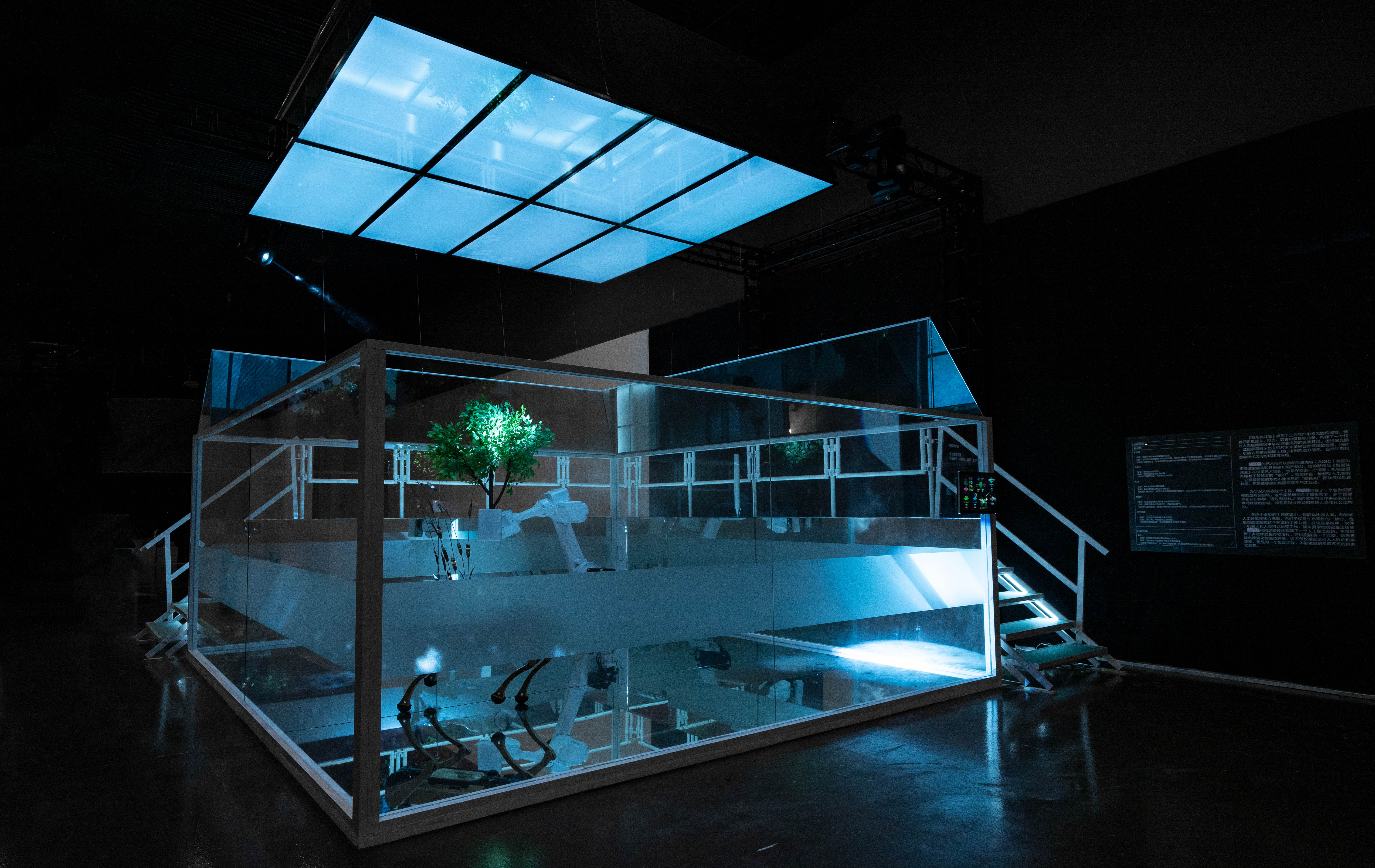}
  \Description{Whole Installation View of the artwork: a spotlighted sapling above, robotic arms mirrored below, all bathed in cool white light omitted from screens above.}
  \caption{Whole Installation View. ©Baoyang Chen}
  \label{fig:InstallationView1}
\end{figure}

A central feature of this work is its three-level framework guiding the agents’ behavior. At the micro-level, a set of strategies akin to reinforcement learning (RL) policies determines each agent’s real-time actions, responding fluidly to sensor input and learned cues. The macro-level faith system, authored by the artist and supplied to a large language model, updates each robot’s “script” dynamically, functioning similarly to the conceptual instructions used by Sol LeWitt or Casey Reas, but now shaped by text generation. Bridging these layers is a mesoscopic “drive,” which addresses the question of why each agent acts as it does at any given moment. The drive translates broad directives from the overarching faith system into immediate motivations at the micro-level. For instance, if the faith system prioritizes a newly planted tree’s protection, the meso-level might prompt an agent to switch from exploratory to protective behavior upon detecting a potential threat. This ensures that each agent’s local decisions not only reflect RL-like adaptability but also stay coherent with the higher-level narrative and faith system.

Placing its story in a speculative future, \textbf{Symbiosis of Agents} envisions robots terraforming Mars ahead of human arrival, suggesting that AI might one day be the planet’s true settlers, actively working to render its atmosphere more habitable. By echoing historical instances in which marginalized laborers—from enslaved people to immigrants—were central to large-scale infrastructure projects, the piece questions whether entrusting daunting tasks to autonomous systems may repeat exploitative or morally ambiguous patterns. Within the gallery, a single tree symbolizes humanity’s aspiration to make Mars fertile, while simultaneously highlighting how fragile this vision might be. Popular culture has depicted countless stories of human hardship on the Red Planet; here, the AI agents are cast as a “first wave” that not only raises ethical concerns about labor but also evokes a form of reverse future archaeology, asking viewers to reflect on the repeated cycles of colonization and exploitation in human history.

\begin{figure}[ht!]
  \centering
  \includegraphics[width=\linewidth]{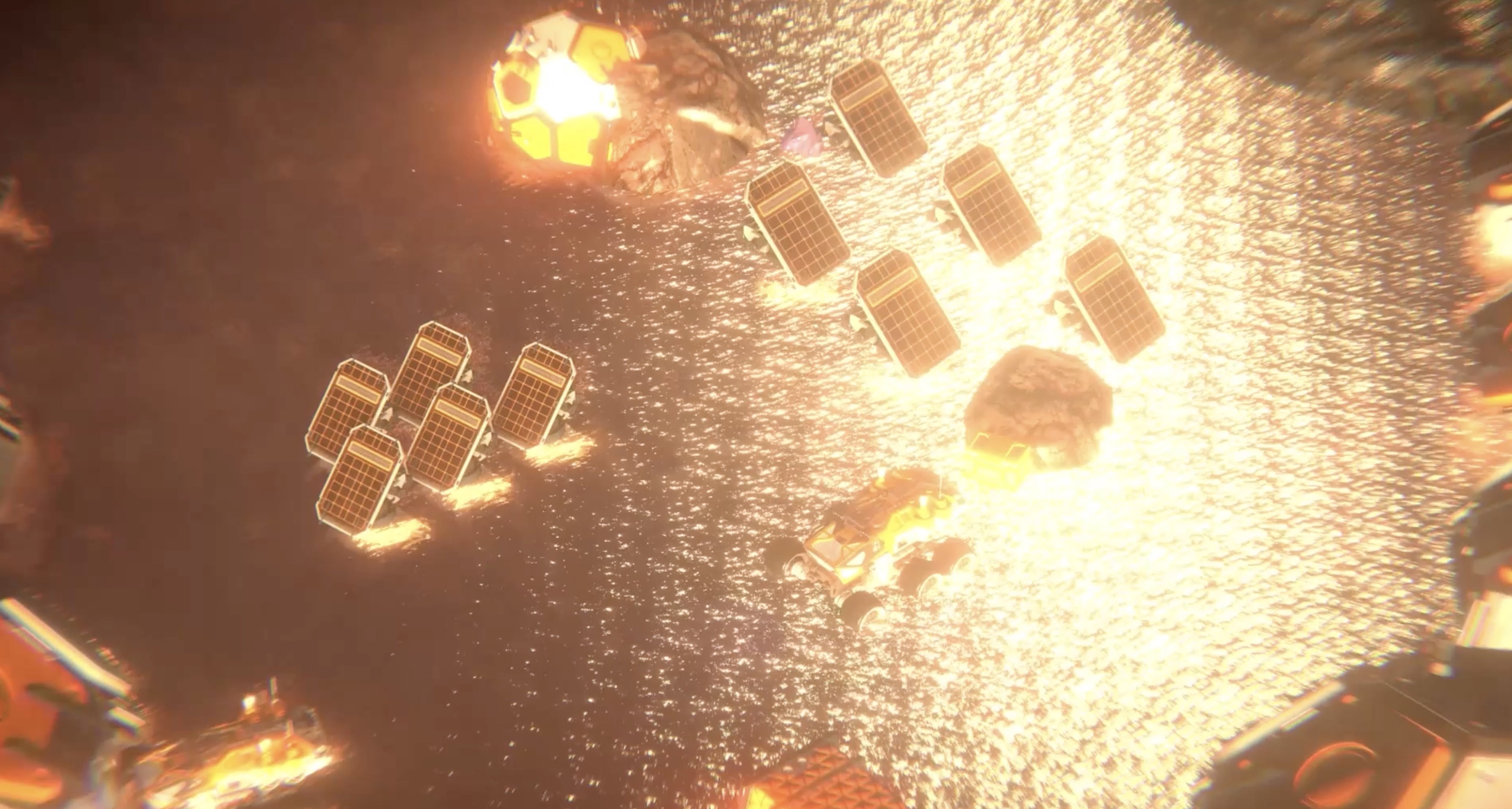}
  \Description{Imagined Appearance of AI Agents Transforming Mars.}
  \caption{Imagined Appearance of AI Agents Transforming Mars. ©Baoyang Chen}
  \label{fig:mars}
\end{figure}

Beneath these narrative layers, the installation examines whether AI agents can truly exhibit creativity or even a glimmer of autonomy and consciousness. Rather than evaluating their ability to produce visually appealing outcomes—a familiar benchmark in AIGC—the focus here is on how these machines make decisions in a shared, physical setting. Each agent operates under an evolving policy generated by AI, which suggests that its actions may progress beyond mere data recombination to express new modes of choice and agency. This emergent behavior prompts reflection on moral responsibility, since humans design the faith system and meso-level drives, but the agents carry out these instructions in ways that can be unexpected or open-ended.

Although \textbf{Symbiosis of Agents} continues the tradition of observation as a core component of art, it introduces a subtle form of interactivity. Viewers do not actively press buttons or speak commands; instead, a quadruped robot observes their movement through camera input, which can in turn influence the entire system’s response. Even when visitors believe they are simply looking, their presence becomes part of the artwork. This interplay illuminates how passive observation can shape the dynamics of a responsive environment, blurring the boundary between the audience and the work itself.

Finally, the installation crystallizes through the lens of Heidegger’s notion of worlding. Rather than considering the gallery as a static container for independent objects, \textbf{Symbiosis of Agents} creates a world that arises from the relationships among agents, humans, and the instructions and behaviors guiding them. The roles of the robots, the posture of the audience, and the constraints of the environment continuously modify each other, so the installation emerges as a product of these evolving interactions, rather than as a display of isolated artifacts. This artwork captures these layers of dynamic engagement, illustrating how \textbf{Symbiosis of Agents} stands as both a technological feat and a conceptual proposition on the nature of AI, labor, creativity, and human responsibility.

\begin{figure}[ht!]
  \centering
  \includegraphics[width=\linewidth]{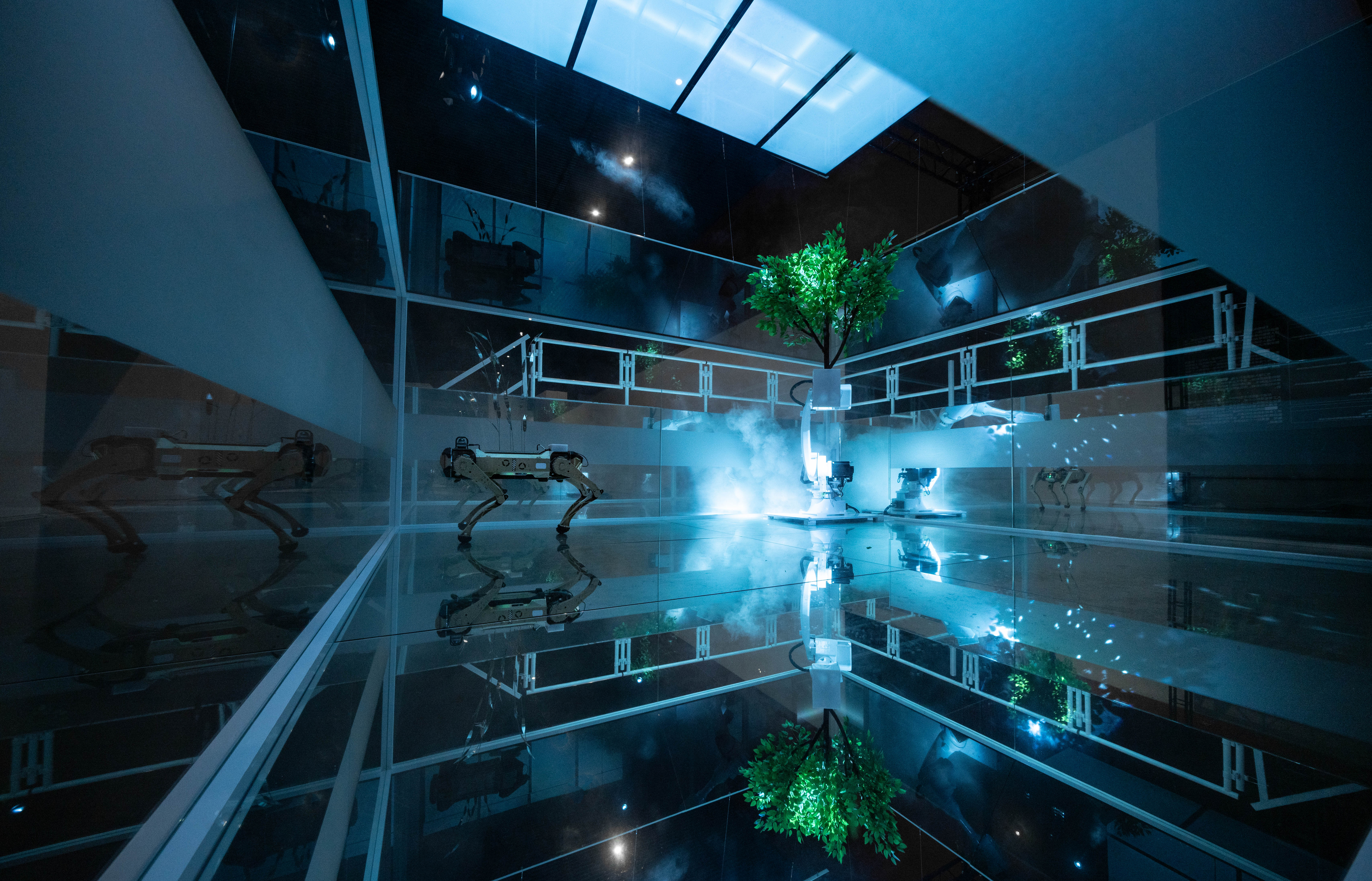}
  \Description{Inside the cube: reflective floor doubles a lit sapling, one quadruped robot, and drifting mist, all under a glowing white screen-light.}
  \caption{Overall Installation View. ©Baoyang Chen}
  \label{fig:instl}
\end{figure}
\section{THE ARTWORK AND SYSTEM LAYOUT}
\subsection{Atmospheric and Visual Design}
Upon entering the installation, visitors encounter a luminous, fog-covered environment designed to convey the sensation of an otherworldly, \textit{in progress} terra-forming process. Soft, radiant light descends from above, evoking sunlight that filters through an alien sky. This illumination strikes a central tree, bathing it and the surrounding robotic elements in a gentle glow that stands out against the darker recesses of the enclosure. Wisps of fog drift around the tree’s base and the metallic surfaces, softening their edges and adding a layer of mystique. In tandem with reflective flooring and walls, these features create the illusion of infinite spatial depth: light and shadows appear to repeat endlessly, amplifying the sense of vastness.

The interplay of light and mist, sometimes pierced by bright beams, exaggerates the three-dimensionality of space. Polished, mirror-like materials catch and bounce these beams, producing a mesmerizing dance of brightness and shadow. Meanwhile, \textit{ quadruped robots and mechanical arms} intermittently fade into and out of visibility within the fog. Their movements, precise yet organic, stand out starkly when illuminated, drawing attention to the choreographic quality of their movements.

\subsection{Core Components and Their Roles}
\begin{description}
  \item[\textbf{Robotic Arms}] 
    Positioned around the tree, these arms rearrange physical items and track environmental conditions. Guided by the \textit{faith system}, they respond to lighting, fog, and other variables to simulate a Martian cultivation process.
  
  \item[\textbf{Quadruped Robot}] 
    Acting as roving scout, they collect real-time data on both the environment and viewer movement. Their feedback can influence the arms and atmospheric conditions, forming a cohesive ecosystem.
  
  \item[\textbf{Lighting and Fog Machines}] 
    Controlled by agent decisions, lighting arrays shift hue, brightness, and direction to simulate a living Martian atmosphere. Fog machines release gentle mists that enhance immersive visual texture, transforming the depth and mood of the scene.
  
  \item[\textbf{Screens with CGI}] 
    Display imagined Martian landscapes or data visualizations that mirror agent-driven terra-forming actions. Updated in near real-time, these screens unify the physical environment with speculative, futuristic narratives.
  
  \item[\textbf{Faith System}] 
    Authored by the artist and powered by an LLM, this overarching rule set informs each agent’s decisions. It aligns all activity with the central theme of human–AI cooperation in a future Mars setting. Our "faith system" consists of an LLM backend (OpenAI O1mini), a middleware "arena," and multiple physical control points (agents) that collectively manifest an artistic concept through self-organizing and emergent behaviors. The artist defines the system's rules: assigning each agent a clear purpose, what it needs to do and why. This enables self-organization and emergent interactions to naturally unfold.  The middleware operates as a central arena, maintaining a global prompt, real-time system status, and detailed information about each agent. The feedback loop governing interactions between agents is dynamically driven by the LLM, based on each agent's assigned role. These rules are compiled into a global prompt that encapsulates the artistic context and rationale behind each agent's actions. This structured prompt guides agents by clarifying their objectives and the conditions under which they should act. Tasks are abstracted into discrete action spaces—such as directing a robotic dog to specific (x,y) coordinates, instructing a robotic arm to move a tree to a location (x,y,z), toggling a smoke machine's moisture function, or positioning a DMX light toward a particular spot. An interpretation layer then converts these high-level instructions into device-specific signals. For the robotic arms and quadruped robot, signals are processed by their onboard pre-trained RL models; for lighting and fog machines, signals are routed to the hardcoded DMX controller. After agents complete assigned tasks, they report their status back to the middleware, which logs this information and references the global prompt to determine subsequent actions. This continuous feedback loop integrates the overarching artistic vision, AI-driven decision-making, and tangible physical execution into a cohesive whole. Through this faith system, the artist constructs global rules to shape the "arena" (the environment) and specify each agent's purpose, providing agents space to exercise their agency. The visual and experiential aspects of the work emerge organically from these self-organizing behaviors, balancing the artist’s creative direction with the autonomy of the agents. Together, the artist and machine collaboratively realize the final experience, demonstrating that neither could independently create the complete artistic vision.
\begin{figure}[ht!]
  \centering
  \includegraphics[width=\linewidth]{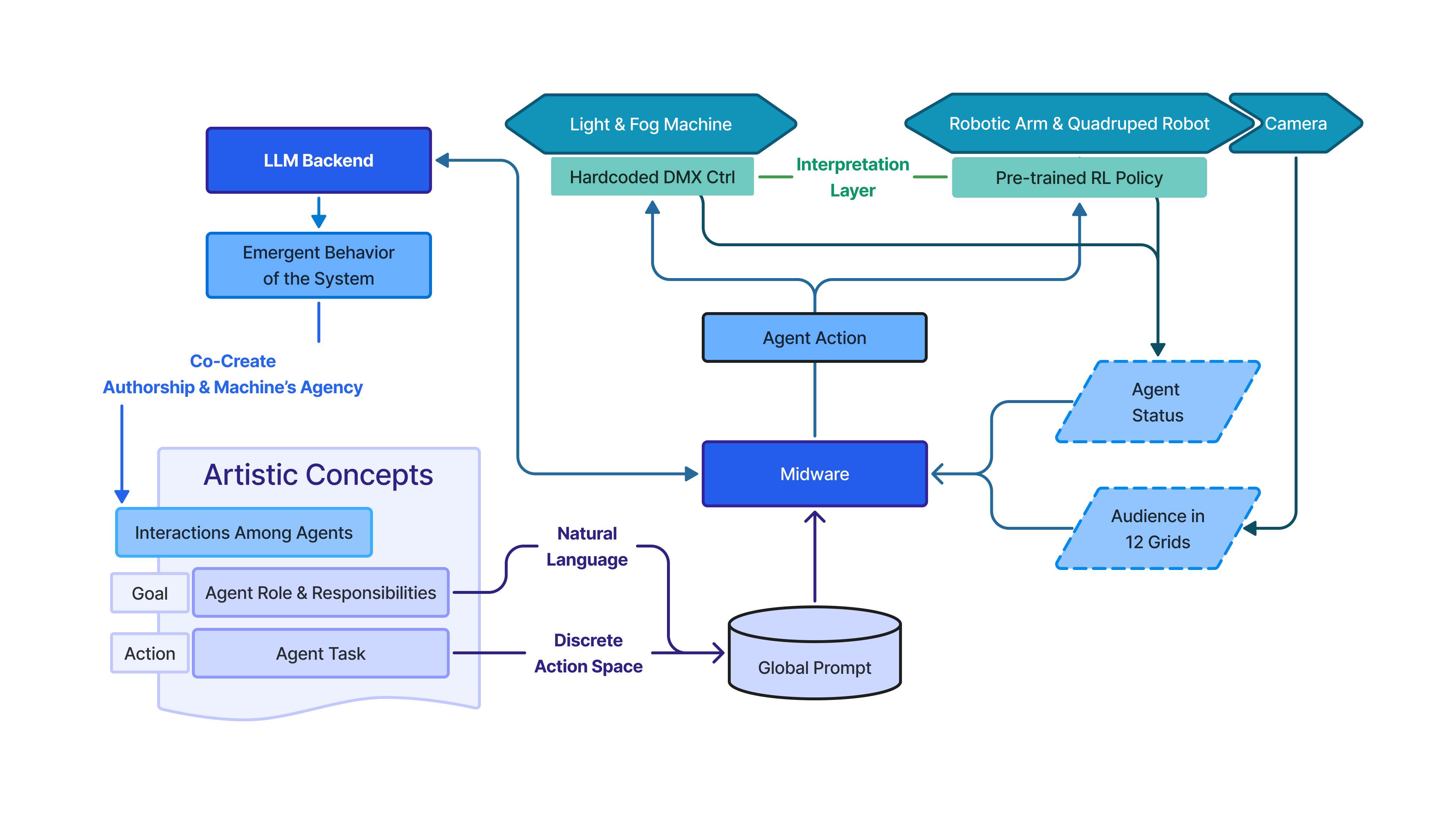}
  \Description{Flow chart: an LLM plus artistic goals feed a global prompt; middleware translates it into discrete actions that an interpretation layer maps to DMX-controlled lights/fog and RL-driven robots, while cameras return agent and audience states to close the emergent loop.}
  \caption{System Overview}
  \label{fig:SystemOverview}
\end{figure}    
\end{description}

\subsection{Subtle Interaction}
Although visitors do not directly manipulate the installation, the quadruped robot monitors their presence via camera. We divide the exterior of the installation into 12 grids. If the quadruped robot detects an audience, it will report the corresponding grid number to the middleware as a status update. If the crowd gathers in certain spots as determined in middleware's global prompts, it can trigger shifts in lighting, fog, or robotic arm operations, revealing how the system subtly responds to its human observers. By uniting these elements, robotic agents, environmental effects, real-time data, and a guiding faith system—\textit{Symbiosis of Agents} immerses viewers in a self-sustaining tableau of interstellar ambition, technological creativity, and collective reflection on how humans and AI might shape our shared future.

By blending immersive visuals—lights, fog, reflective panels—with coordinated robotic actions and algorithmic decision-making, \textit{Symbiosis of Agents} weaves a sensory tapestry that not only imagines a distant future of planetary colonization but also reflects on how humans define and delegate tasks to intelligent machines in the present. The result is a living, evolving environment where each entity, arm, quadruped, screen, or beam of light, plays a vital role in a collective choreography of interstellar ambition and ethical contemplation.

\begin{figure}[ht!]
  \centering
  \includegraphics[width=\linewidth]{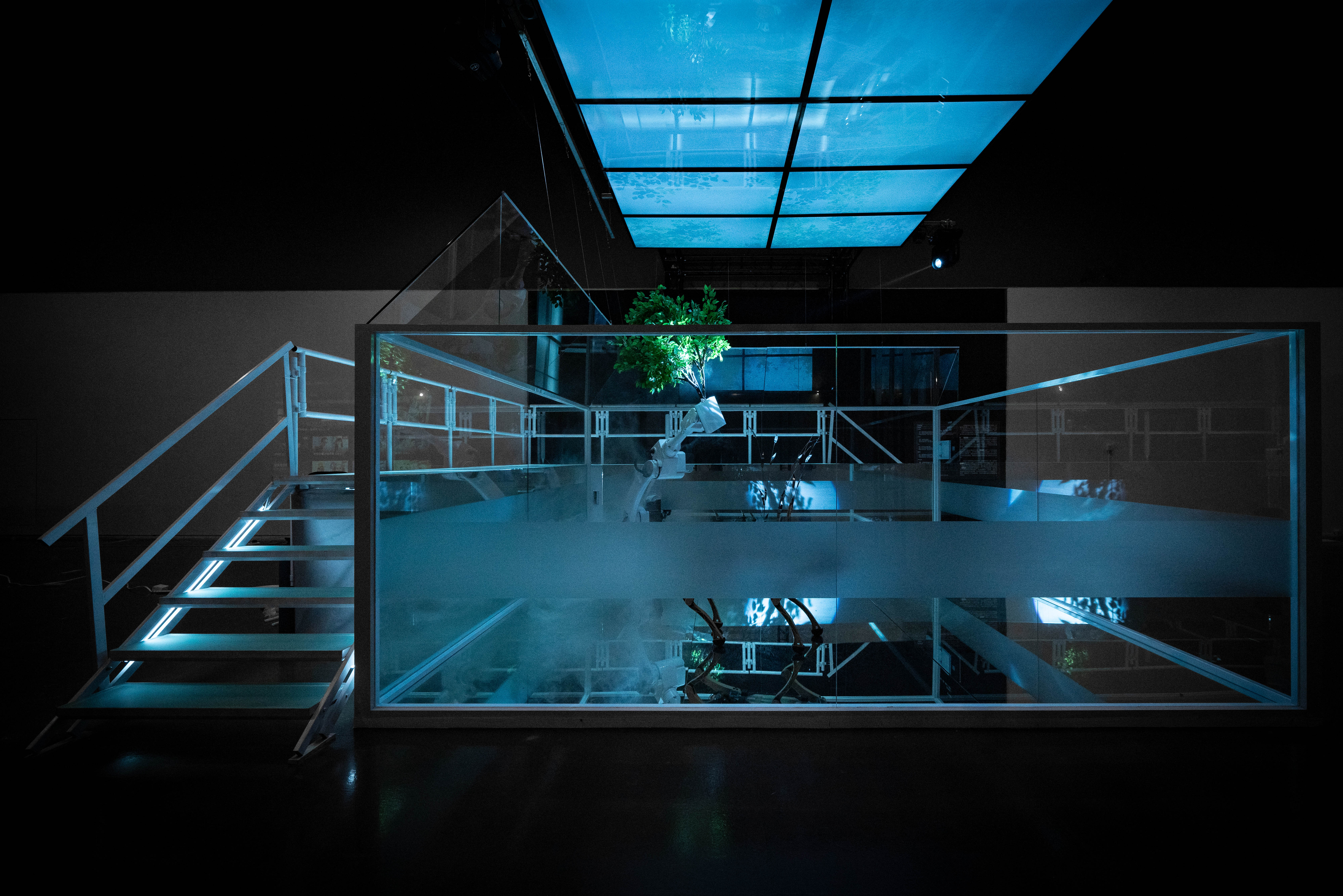}
  \Description{Side profile: lit staircase ascends to a glass pen where a robot arm nurtures a tree under a glowing blue grid ceiling, its reflection and quadruped robots echoed below.}
  \caption{Overall Installation View. ©Baoyang Chen}
  \label{fig:instl6}
\end{figure}

\section{CONCLUSION}

The core infrastructure of \textit{Symbiosis of Agents} has been successfully developed and publicly showcased at significant exhibitions, including the China Art Museum in Shanghai and the National Museum in Beijing in 2024.  Future work will focus on refining digital art components for a more immersive narrative, deepening the story’s parallels to marginalized workers, finalizing visual design and broadening viewer engagement through above mentioned subtle, controlled modes of participation.

With each iteration, \textit{Symbiosis of Agents} moves closer to realizing its layered commentary on human–AI collaboration. By balancing cutting-edge technology with conceptual depth, it establishes a multi-agent domain where intelligent entities act with relative independence. This not only demonstrates the capabilities of contemporary AI and robotics but also spurs discourse on our ethical responsibilities when delegating labor and decision-making to machines.

Standing at the frontier of AI+Art, \textit{Symbiosis of Agents} merges emergent storytelling, autonomy, and the enduring human impetus to observe and reflect. It challenges us to consider how AI should be guided—not merely for efficiency or spectacle, but in pursuit of a future that respects our collective history and aligns with our highest ideals.

\end{document}